\documentclass[letterpaper,10pt,journal,twoside]{IEEEtran}
\usepackage{amsmath, amssymb, amsfonts}
\usepackage{algorithmic}
\usepackage{graphicx}
\usepackage{textcomp}
\usepackage{xcolor}
\usepackage{cite}
\usepackage{mathtools} 
\usepackage{amsthm}      
\usepackage[colorlinks=true, linkcolor=blue, citecolor=red]{hyperref} %
\usepackage{orcidlink}
\usepackage{epstopdf}   
\usepackage{booktabs}   

\usepackage{tabularx}
\usepackage{multirow}
\usepackage{subfig}

\usepackage{tabularx}
\newcolumntype{C}{>{\centering\arraybackslash}X} 

\newcommand{\vect}[1]{\boldsymbol{#1}}
\newcommand{\mat}[1]{\mathbf{#1}}

\newtheoremstyle{myremarkstyle}
  {\topsep}{\topsep}{\itshape}{}{\bfseries}{:}{.5em}{}
\theoremstyle{myremarkstyle}

\theoremstyle{definition}
\newtheorem{definition}{Definition}[section]
\newtheorem{assumption}{Assumption}[section]
\theoremstyle{plain}
\newtheorem{lemma}{Lemma}[section]
\theoremstyle{remark}
\newtheorem{remark}{Remark}[section]

\title{Appointed-Time Fault-Tolerant Fast Terminal Sliding Control for UAVs with a Tracking Error-Based Disturbance Observer}

\author{Jinfa Li~\orcidlink{0009-0004-9819-7661}, Sheng Xu*~\orcidlink{0000-0002-6314-8104}, Yanjie Wu, Chengyue Su
\thanks{This work was supported in part by the National Natural Science Foundation of China (Grant No. 32271996).(Corresponding author: Sheng Xu).}
\thanks{Jinfa Li, Sheng Xu are with the Unmanned Systems Technology Laboratory, also with the School of Physics and Optoelectronic Engineering, Guangdong University of Technology, Guangzhou 510006, China (e-mail: 2112315048@mail2.gdut.edu.cn, xusheng@gdut.edu.cn). }
\thanks{Yanjie Wu is with the Guangdong Provincial Key Laboratory of Sensing Physics and System Integration Applications, Guangdong University of Technology, Guangzhou 510006, China (e-mail: wuyanjie@gdut.edu.cn).}
\thanks{Chengyue Su is with the School of Advanced Manufacturing, Guangdong University of Technology, JieYang 515200, China (e-mail: cysu@gdut.edu.cn).}       
}

\begin{document}

\maketitle

\begin{abstract}

This paper addresses the critical challenge of high-performance fault-tolerant tracking control for quadrotor UAVs under complex conditions including external disturbances, parameter uncertainties, and actuator faults. A synergistic control framework  is proposed by decoupling the system into position and attitude loops, each governed by the following three innovations. First, an Appointed-Time Prescribed Performance Function (AT-PPF) is developed to strictly constrain the tracking error within a predefined funnel boundary, ensuring convergence within a fixed time regardless of initial conditions. Second, a Tracking Error-Based Fixed-Time Composite Disturbance Observer (TFxTCDO) is developed to enable fast and accurate estimation of lumped disturbances without requiring full-state information or precise system modeling. This observer directly couples disturbance estimation with the control objective, thereby enhancing robustness and convergence. Third, a  Fast Second-Order Terminal Sliding Mode Controller (NFSO-TSMC) is designed to achieve faster convergence  and effective chattering mitigation. Simulation results demonstrate that the proposed control strategy significantly outperforms conventional methods in terms of tracking accuracy, fault recovery speed, and robustness against uncertainties.

\end{abstract}


\section{Introduction}


\IEEEPARstart{T}he low-altitude economy is rapidly emerging as a strategic emerging sector with tremendous potential. \cite{intro_low_altitude_economy_potential}. Within the low-altitude economy platform, quadrotor Unmanned Aerial Vehicles (UAVs) have found the most widespread application due to their excellent maneuverability and broad application prospects \cite{intro_uav_cornerstone}. However, as the application scenarios expand, quadrotor UAVs are increasingly confronted with complex operational environments. A host of challenges, including external disturbances, parameter uncertainties, input saturation, and actuator faults, invariably arise, posing significant threats to flight stability and mission reliability. For instance, dense urban environments characterized by high-rise buildings can generate complex turbulence, subjecting the UAV to severe and unpredictable disturbances. Prolonged operation may lead to the gradual performance degradation of critical components, thereby amplifying parameter uncertainties. Furthermore, during high-payload transportation or aggressive maneuvers, the issue of input saturation frequently occurs, constraining the vehicle's control authority. More critically, the occurrence of an actuator fault directly compromises flight safety and mission execution. 

Therefore, investigating fault-tolerant tracking control with prescribed performance for quadrotor UAVs operating in complex environments is of paramount importance. Such research holds significant theoretical value in advancing control theory and, more pressingly, addresses an urgent practical need to ensure the safe and efficient development of the low-altitude economy. This paper is dedicated to tackling these challenges by proposing a robust and unified control framework.

In the domain of Unmanned Aerial Vehicle (UAV) control, Prescribed Performance Control (PPC) has been widely adopted as a powerful technique to enhance system robustness and convergence speed. A significant body of research has evolved around the design and application of various types of Prescribed Performance Functions (PPFs), prominently including the Finite-Time Prescribed Performance Function (FTPPF), the Fixed-Time Prescribed Performance Function (FxTPPF), and the Appointed-Time Prescribed Performance Function (AxTPPF). The FTPPF method ensures that the system converges within a finite time, with the convergence duration being dependent on the initial conditions of the system. A study successfully employed an FTPPF to achieve rapid trajectory tracking for a UAV\cite{ppc_ftppf_example}. However, a notable limitation of this approach is that the convergence time remains susceptible to the initial error magnitude; a large initial error can lead to a comparatively slower convergence process. To overcome this dependency, the FxTPPF method was developed, which guarantees system convergence within a fixed-time upper bound, irrespective of the initial conditions. The work in \cite{ppc_fxtppf_example} utilized a fixed-time control strategy, demonstrating improved rapid response capabilities and enhanced resilience against external disturbances in UAVs. Nevertheless, under conditions of high nonlinearity and strong uncertainties, the convergence rate of FxTPPF might degrade, and it can potentially demand excessive control energy. More recently, the AxTPPF has emerged as a compelling alternative, offering the distinct advantage of allowing the user to pre-specify the convergence time, which is independent of both initial conditions and system parameters. This feature renders AxTPPF particularly advantageous for tasks with stringent temporal requirements, such as navigation in dynamic environments or cooperative control, when compared to FTPPF and FxTPPF. Recent studies have substantiated the high efficiency of the AxTPPF methodology in managing uncertain systems \cite{ppc_axtppf_example}. In the context of complex low-altitude environments, where quadrotor UAVs are subjected to a confluence of challenges including external disturbances, input saturation, and actuator faults, conventional finite-time or fixed-time approaches may struggle to meet the demanding requirements for high timeliness and strict performance constraints. Therefore, this paper adopts an Appointed-Time Prescribed Performance Function (AxTPPF). By leveraging AxTPPF, we ensure that the tracking error converges within a pre-assigned time interval, independent of the initial state. 

Sliding Mode Control (SMC) has garnered significant research interest for quadrotor applications. Depending on the design of the sliding surface and the control law, SMC methodologies can be categorized into several prominent variants, including Terminal Sliding Mode Control (TSMC), Finite-Time Sliding Mode Control (FTSMC), Fixed-Time Sliding Mode Control (FxTSMC), Integral Sliding Mode Control (ISMC), and Recursive Sliding Mode Control (RSMC). Terminal Sliding Mode Control (TSMC) is designed to drive system states to the equilibrium in finite time. TSMC is proposed for quadrotor trajectory tracking\cite{smc_tsmc_example}, which successfully enhanced the convergence rate. However, a persistent challenge with conventional TSMC is the pronounced chattering phenomenon, which can compromise the execution of high-precision tasks \cite{smc_chattering_issue}. To address this, Finite-Time Sliding Mode Control (FTSMC) was developed, offering robust performance against disturbances while ensuring finite-time convergence\cite{smc_ftsmc_example}. Nonetheless, the convergence time of FTSMC is contingent upon the initial system states, rendering it less suitable for missions with strict temporal constraints\cite{smc_ftsmc_limitation}. Subsequently, Fixed-Time Sliding Mode Control (FxTSMC) emerged, which, through the construction of an appropriate control law, guarantees convergence within a fixed time independent of the initial conditions \cite{smc_fxtsmc_example}. However, a notable drawback is its potentially high computational complexity, particularly when implemented in higher-order or strongly nonlinear systems \cite{smc_fxtsmc_limitation}. On another front, Integral Sliding Mode Control (ISMC) enhances robustness against external disturbances by incorporating an integral term into the sliding surface. Its application to UAV attitude control, as demonstrated in \cite{smc_ismc_example}, has shown superior disturbance rejection capabilities. However, a possible limitation of this approach is the introduction of additional dynamic errors due to the integral action, which may compromise control precision and degrade overall performance \cite{smc_ismc_limitation}. Lastly, Recursive Sliding Mode Control (RSMC) employs a multi-layered, nested sliding surface design to achieve recursive-like approximation control for complex systems\cite{smc_rsmc_example}. This method, however, is often computationally intensive and relies heavily on high-precision state observation \cite{smc_rsmc_limitation}. Addressing the limitations of the aforementioned methods, this paper presents a \textbf{Fast Second-Order Terminal Sliding Mode Control (NFSO-TSMC)}. This innovative approach synergizes the chattering-suppression characteristics of second-order SMC with the efficient convergence properties of TSMC. The proposed NFSO-TSMC ensures global convergence within a prescribed time while simultaneously mitigating chattering, thereby enhancing both the tracking precision and dynamic response performance of the quadrotor in complex environments.To overcome these challenges, we put forth a Fast Second-Order Terminal Sliding Mode Control (NFSO-TSMC), which artfully blends the merits of second-order and terminal SMC to refine tracking precision and dynamic responsiveness. 

To ensure the stable and fault-tolerant control of quadrotor UAVs in complex environments, researchers have developed a variety of Fault-Tolerant Control (FTC) strategies to counteract issues such as actuator faults, sensor failures, and external disturbances. FTC approaches can be broadly classified into Passive FTC, Active FTC, Reconfigurable FTC, Data-Driven FTC, and Hybrid FTC \cite{ftc_overview}. Passive FTC methods are designed to be inherently robust by considering system uncertainties and potential fault effects during the initial controller design phase. This allows the controller to maintain system stability upon fault occurrence without requiring online parameter adjustments. For instance, robust techniques like Sliding Mode Control (SMC) \cite{ftc_passive_smc} and H-infinity control \cite{ftc_passive_hinf} have been employed to enhance a UAV's resilience to actuator faults. However, the adaptability of passive methods is limited; they may still suffer from performance degradation when faced with severe actuator faults or significant variations in system parameters. In contrast, Active FTC relies on a Fault Detection and Diagnosis (FDD) module to dynamically compensate for fault effects by adjusting the control law in real-time. Adaptive Control \cite{ftc_active_adaptive} and Model Reference Adaptive Control (MRAC) \cite{ftc_active_mrac} are widely utilized within this paradigm\cite{ftc_active_adaptive_example}, thereby improving flight stability. A drawback of adaptive control, however, is the potential for estimation delays of system parameters, which can affect the response speed. Model Predictive Control (MPC) has also been applied to UAV fault tolerance \cite{ftc_active_mpc}, offering the ability to optimize control inputs by considering future states, but its high computational complexity often hinders real-time implementation. Reconfigurable FTC achieves fault tolerance through control law adaptation or Control Allocation (CA), as demonstrated in \cite{ftc_reconfigurable_ca}, where stable flight is maintained despite the complete failure of a single actuator. Nevertheless, such methods typically necessitate redundant actuators or sophisticated thrust allocation schemes, which increases system complexity and hardware costs. More recently, data-driven approaches have been introduced to the FTC landscape, leveraging Reinforcement Learning (RL) \cite{ftc_datadriven_rl} and Deep Neural Networks (DNNs) \cite{ftc_datadriven_dnn} for fault compensation, enabling UAV to maintain stability under unknown fault conditions\cite{ftc_datadriven_drl_example}. The performance of these methods, however, is highly dependent on the quality of the training data, and their generalization capabilities require further enhancement. In comparing these paradigms, Active FTC demonstrates superior adaptability and real-time performance by promptly detecting and compensating for faults without relying on hardware redundancy or large datasets, making it well-suited for dynamic, real-time applications\cite{ftc_active_advantages}.

Disturbance observers have been widely employed in control systems to enhance robustness by estimating and compensating for external uncertainties in real time. Among them, state error-based disturbance observers are the most commonly used, wherein the estimation law is formulated based on the deviation between actual system states and their model-based predictions. These observers have been successfully applied in numerous scenarios, providing reliable disturbance estimation under nominal conditions.
However, such state error-based schemes typically rely on accurate system modeling and full-state availability~\cite{Course2023}. In practical applications---particularly in highly nonlinear or partially observable systems~\cite{Dabush2023}---model mismatches, unmeasured states, or sensor limitations may significantly degrade the estimation accuracy.
Moreover, the estimation dynamics are often decoupled from the control objective, leading to slower convergence or reduced fault response speed in safety-critical scenarios.
To overcome these limitations, we propose a \textbf{Tracking Error-Based fixed-time composite disturbance observer}(TFxTCDO). Unlike conventional approaches, our observer leverages the tracking error, which inherently reflects the deviation from the control objective, thereby enabling tighter coupling between disturbance estimation and performance requirements. This design not only improves convergence rate and estimation precision but also allows for model-free deployment under fixed-time convergence guarantees, making it particularly suitable for real-time fault-tolerant control of quadrotor UAVs operating in uncertain and complex environments.

In light of the preceding analysis, this paper introduces a framework for ensuring fast, robust, and fault-resilient trajectory tracking under complex operating conditions. This is achieved through a synergistic design comprising: a new appointed-time prescribed performance function, a  fixed-time sliding mode surface, and an tracking error-based disturbance observer that incorporates both actuator fault detection and disturbance compensation.  
The main contributions of this study are summarized as follows:
\begin{enumerate}
    \item \textbf{A Prescribed-Time Fault-Tolerant Tracking Control Scheme:} We have designed a comprehensive control scheme for both the position and attitude subsystems. This scheme guarantees stable tracking control even in the presence of actuator faults, with convergence achieved within a user-prescribed time that is independent of the initial conditions.
    
    \item \textbf{A Tracking Error-Based Fixed-Time Composite Disturbance Observer:}  A fixed-time disturbance observer based on tracking error is proposed, enabling fast, model-independent estimation of composite disturbances while tightly integrating with control objectives to enhance robustness and convergence.
    
    \item \textbf{A  Fast Second-Order Terminal Sliding Mode Control (NFSO-TSMC) Method:} We have designed a  NFSO-TSMC that effectively improves both the transient and steady-state response of the system. This method enhances control precision and response speed while substantially mitigating the chattering phenomenon inherent in conventional sliding mode control.
\end{enumerate}


The remainder of this paper is organized as follows. Section~\ref{sec:problem_formulation} presents the mathematical model of the quadrotor UAV and the necessary preliminaries. Section~\ref{sec:controller_and_observer} details the design of the controller and the disturbance observer. Section~\ref{sec:simulations} provides simulation results to validate the proposed control strategy. Finally, Section~\ref{sec:conclusion} concludes the paper and outlines the future research.

\section{Problem Formulation and Preliminaries}
\label{sec:problem_formulation}
This section establishes the mathematical foundation for the control problem. We begin by presenting the dynamic model of the quadrotor UAV, followed by the necessary preliminaries for the subsequent control design.

\subsection{Quadrotor Dynamic Model}
\label{sec:dynamics}
\begin{figure}[h!] 
    \centering
    \includegraphics[width=0.6\textwidth]{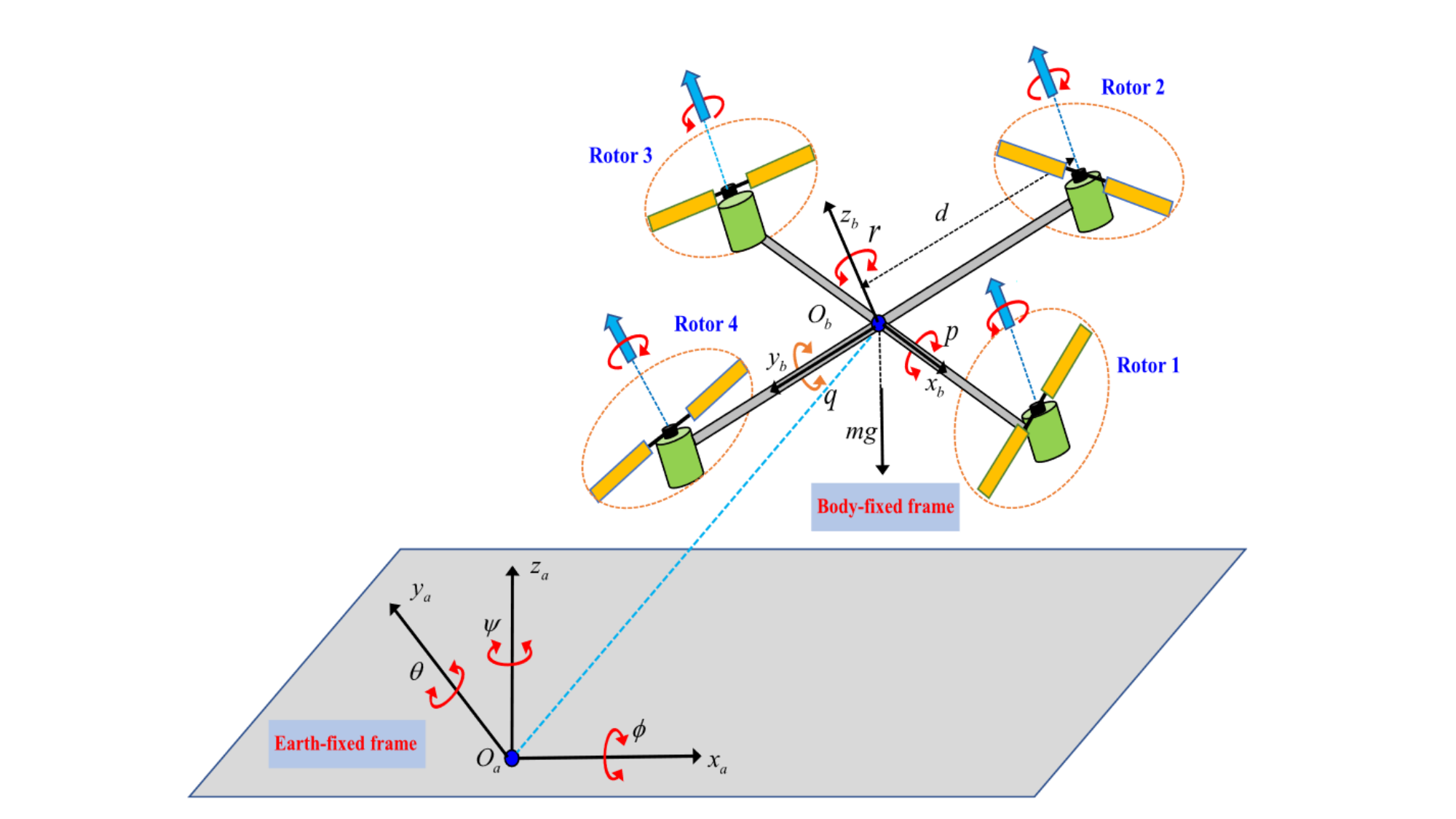} 
    \caption{Schematic of the quadrotor UAV model. The figure illustrates the 'X' configuration, the body frame ($O_B-X_B-Y_B-Z_B$) relative to the inertial North-East-Down (NED) frame ($O_I-X_I-Y_I-Z_I$), the rotor numbering and rotation directions, and the corresponding thrust and control torques.}
    \label{fig:quad_model} 
\end{figure}

The development of the quadrotor's dynamic model is based on the following standard assumptions. A schematic representation of the quadrotor model, including the coordinate frames and control inputs, is provided in Figure~\ref{fig:quad_model}.

\begin{assumption}
The North-East-Down (NED) coordinate system is adopted as the inertial reference frame. The effects of Earth's rotation and curvature are considered negligible.
\end{assumption}
\begin{assumption}
The quadrotor is modeled as a rigid body with a perfectly symmetric structure, where its center of mass (CoM) coincides with the geometric center.
\end{assumption}
\begin{assumption}
The thrust and drag forces, as well as the torques generated by the propellers, are proportional to the square of their rotational speeds.
\end{assumption}

Based on these assumptions, the 6-DOF dynamic model of the quadrotor is given by:

\subsubsection{Translational Dynamics}
\begin{align}
\ddot{x} &= \frac{T_1}{m}\left(\cos\phi\sin\theta\cos\psi + \sin\phi\sin\psi\right) - \frac{c_x}{m}\dot{x} + \bar{d}_x \label{eq:trans_x} \\
\ddot{y} &= \frac{T_1}{m}\left(\cos\phi\sin\theta\sin\psi - \sin\phi\cos\psi\right) - \frac{c_y}{m}\dot{y} + \bar{d}_y  \\
\ddot{z} &= \frac{T_1}{m}\cos\phi\cos\theta - g - \frac{c_z}{m}\dot{z} + \bar{d}_z \label{eq:dyn_z}
\end{align}

\subsubsection{Rotational Dynamics}
\begin{align}
\ddot{\phi} &= \frac{T_2}{I_{xx}} + \dot{\theta}\dot{\psi}\frac{I_{yy}-I_{zz}}{I_{xx}} - \frac{c_\phi}{I_{xx}}\dot{\phi} + \bar{d}_\phi \label{eq:dyn_phi} \\
\ddot{\theta} &= \frac{T_3}{I_{yy}} + \dot{\phi}\dot{\psi}\frac{I_{zz}-I_{xx}}{I_{yy}} - \frac{c_\theta}{I_{yy}}\dot{\theta} + \bar{d}_\theta \label{eq:dyn_theta} \\
\ddot{\psi} &= \frac{T_4}{I_{zz}} + \dot{\phi}\dot{\theta}\frac{I_{xx}-I_{yy}}{I_{zz}} - \frac{c_\psi}{I_{zz}}\dot{\psi} + \bar{d}_\psi \label{eq:rot_psi}
\end{align}
where $(x, y, z)$ are the positions, and $(\phi, \theta, \psi)$ are the Euler angles. The terms $\bar{d}_{(\cdot)}$ represent lumped disturbances. The quadrotor dynamics can be unified into a compact form:
\begin{equation}
\ddot{\nu}_i = U_i + F_i + D_i, \quad i = 1, \dots, 6
\label{eq:unified_dyn}
\end{equation}
where $\vect{\nu} = [x, y, z, \phi, \theta, \psi]^T$, and $U_i, F_i, D_i$ consolidate the control inputs, known dynamics, and disturbances. To achieve position control, the desired virtual control inputs for translation are converted into reference attitude angles:
\begin{align}
\nu_4^d &= \arcsin\left(\frac{U_1\sin(\nu_6^d) - U_2\cos(\nu_6^d)}{\sqrt{U_1^2 + U_2^2 + U_3^2}}\right) \label{eq:inv_kin_phi} \\
\nu_5^d &= \arctan\left(\frac{U_1\cos(\nu_6^d) + U_2\sin(\nu_6^d)}{U_3}\right) \label{eq:inv_kin_theta}
\end{align}

\subsection{Actuator Fault Model}
\label{sec:fault_model}
For a quadrotor in an 'X' configuration, the relationship between the total control input vector $\mat{U} = [U_1, \dots, U_4]^T$ and the vector of squared angular velocities $\vect{\Omega} = [\omega_1^2, \dots, \omega_4^2]^T$ is described by the control allocation matrix, $\mat{B}$:
\begin{equation}
\mat{U} = \mat{B}\vect{\Omega}, \quad \text{where} \quad
\mat{B} = 
\begin{bmatrix}
b & b & b & b \\
\frac{bl}{\sqrt{2}} & -\frac{bl}{\sqrt{2}} & -\frac{bl}{\sqrt{2}} & \frac{bl}{\sqrt{2}} \\
\frac{bl}{\sqrt{2}} & \frac{bl}{\sqrt{2}} & -\frac{bl}{\sqrt{2}} & -\frac{bl}{\sqrt{2}} \\
-d & d & -d & d
\end{bmatrix}
\end{equation}
When considering faults, the actual output $\vect{\Omega_f}$ can be modeled as:
\begin{equation}
\vect{\Omega_f} = (\mat{I} + \vect{\Delta}_m)\vect{\Omega} + \vect{\Delta}_a
\label{eq:fault_model}
\end{equation}
where $\vect{\Delta}_m = \text{diag}(\delta_1, \dots, \delta_4)$ is the multiplicative fault matrix, and $\vect{\Delta}_a = [\epsilon_1, \dots, \epsilon_4]^T$ is the additive fault vector. The actual control input applied to the system thus becomes:
\begin{equation}
\mat{U}_f = \mat{B}\vect{\Omega_f} = \mat{U} + \mat{B}\vect{\Delta}_m\vect{\Omega} + \mat{B}\vect{\Delta}_a
\end{equation}
By defining an extended disturbance term $\mathcal{D}_i = D_i + ( \mat{B}\vect{\Delta}_m\vect{\Omega} + \mat{B}\vect{\Delta}_a )_i$, the system dynamics, considering actuator faults, can be simplified to:
\begin{equation}
\ddot{\nu}_i = U_i + F_i + \mathcal{D}_i
\label{eq:faulty_dyn}
\end{equation}
This formulation preserves the structural consistency with the fault-free model from Eq.~\eqref{eq:unified_dyn}.

\subsection{Prescribed-Time Prescribed Performance Control}
\label{sec:ppc}
The core objective of PTPPC is to ensure the tracking error $e_i(t)$ remains strictly within a time-varying boundary:
\begin{equation}
-\delta_i(t) < e_i(t) < \delta_i(t)
\end{equation}
where $\delta_i(t)$ is the Prescribed-Time Performance Function (PTPF). In this study, we adopt a PTPF that ensures prescribed-time convergence, formulated as:
\begin{equation}
\delta_i(t) =
\begin{cases}
    (\delta_{i0} - \delta_{iT}) \exp\left(-\frac{l_i t}{T_i - t}\right) + \delta_{iT}, & 0 \le t < T_i \\
    \delta_{iT}, & t \ge T_i
\end{cases}
\label{eq:ptpf}
\end{equation}
where $\delta_{i0}$ is the initial bound, $\delta_{iT}$ is the final steady-state bound, and $T_i$ is the user-defined convergence time.

\subsection{Preliminaries and Lemmas}
\label{sec:lemmas}
This subsection presents the key definitions and lemmas fundamental to the stability analysis.
\begin{definition}[Finite-Time Stability \cite{def_finite_time_stability}]
The origin of a system $\dot{x} = f(x)$ is said to be globally finite-time stable if it is Lyapunov stable and its settling-time function $T(x_0)$ is finite for any initial condition $x_0$.
\end{definition}

\begin{definition}[Fixed-Time Stability \cite{def_fixed_time_stability}]
    \label{def_fixed_time_stability} 
    The origin is globally fixed-time stable if it is globally finite-time stable and the settling-time function $T(x_0)$ is bounded by a constant $T_{\max}$ that is independent of the initial conditions.
\end{definition}

\begin{lemma}[Fixed-Time Stability Criterion \cite{lemma_fixed_time_stability}]
If there exists a continuous, positive definite, and radially unbounded function $V(x)$ such that its time derivative satisfies:
\begin{equation}
\dot{V}(x) \le -aV(x)^\gamma - bV(x)^\alpha
\label{eq:lemma_fxt}
\end{equation}
for constants $a > 0$, $b > 0$, $0 < \gamma < 1$, and $\alpha > 1$, then the origin is globally fixed-time stable. The settling time is bounded by $T_{\max} = \frac{1}{a(1-\gamma)} + \frac{1}{b(\alpha-1)}$.
\end{lemma}

\begin{lemma}[Inequality for Sums of Powers \cite{lemma_inequality_sum_powers}]
For any $x_1, \dots, x_n \in \mathbb{R}$ and any $l > 0$, the following inequality holds:
\begin{equation}
\left( \sum_{i=1}^{n} |x_i| \right)^l \le \max\{n^{l-1}, 1\} \sum_{i=1}^{n} |x_i|^l
\label{eq:lemma_inequality}
\end{equation}
\end{lemma}

\section{Controller and Observer Design}
\label{sec:controller_and_observer}

\subsection{Fast Second-Order Terminal Sliding Mode Control}
To address the challenges posed by the quadrotor UAV's inherently nonlinear, strongly coupled dynamics, which are highly susceptible to external disturbances, this section introduces a Fast Second-Order Terminal Sliding Mode Control (FTSMC) strategy. This method synergistically integrates the fixed-time convergence advantage of the fast terminal sliding mode with a robust control design, ensuring strong suppression of disturbances. Consequently, it provides an effective framework for achieving high-precision position and attitude tracking control. The subsequent discussion will systematically elaborate on the sliding surface construction, control law design, and fixed-time convergence analysis.

The derivation begins with the tracking error, defined as $e_i(t) = \nu_i(t) - \nu_i^d(t)$. Taking the second time derivative of the error yields:
\begin{equation}
\ddot{e}_i(t) = \ddot{\nu}_i(t) - \ddot{\nu}_i^d(t)
\end{equation}
Substituting the system dynamics from \eqref{eq:faulty_dyn} for $\ddot{\nu}_i(t)$, we get:
\begin{equation}
\ddot{e}_i(t) = \left( U_{i}(t) + F_i(\nu_i, \dot{\nu}_i, t) + D_i(t) \right) - \ddot{\nu}_i^d(t)
\end{equation}
By rearranging the terms, the error dynamic equation is obtained as:
\begin{equation}
\ddot{e}_i(t) = \left( F_i(\nu_i, \dot{\nu}_i, t) - \ddot{\nu}_i^d(t) \right) + U_{i}(t) + D_i(t)
\end{equation}
where the terms are defined as:
\begin{itemize}
    \item $\mathcal{F}_i = F_i(e_i + \nu_i^d, \dot{e}_i + \dot{\nu}_i^d, t) - \ddot{\nu}_i^d(t)$ represents the known (or estimable) part of the system dynamics.
    \item $\mathcal{D}_i(t) = D_i(t)$ denotes the lumped disturbance.
    \item $U_{i}(t)$ is the actual control input to be designed.
\end{itemize}

To enforce the prescribed performance constraint, $|e_i(t)| < \delta_i(t)$, an error transformation is introduced, defining the transformed error $\xi_i(t)$ as:
\begin{equation}
\xi_i(t) = \text{arctanh}\left(\frac{e_i(t)}{\delta_i(t)}\right)
\label{eq:transformed_error_full}
\end{equation}
The first and second derivatives of $\xi_i(t)$ are given by, respectively:
\begin{equation}
\dot{\xi}_i(t) = \frac{\dot{e}_i(t)\delta_i(t) - e_i(t)\dot{\delta}_i(t)}{\delta_i(t)^2 - e_i(t)^2}
\label{eq:nu_dot_full}
\end{equation}
\begin{multline}
\ddot{\xi}_i(t) = \frac{1}{(\delta_i^2 - e_i^2)^2} \Big[ (\delta_i\ddot{e}_i - e_i\ddot{\delta}_i)(\delta_i^2 - e_i^2) \\
- (\delta_i\dot{e}_i - e_i\dot{\delta}_i)(2\delta_i\dot{\delta}_i - 2e_i\dot{e}_i) \Big]
\label{eq:nu_ddot_full}
\end{multline}
By substituting the error dynamics, the dynamics of the transformed error, $\ddot{\xi}_i(t)$, can be expressed as:
\begin{equation}
\ddot{\xi}_i(t) = \tilde{F}_i + \tilde{G}_i U_{i}(t) + \tilde{D}_i
\label{eq:transformed_dynamics_full}
\end{equation}
where:
\begin{multline}
\tilde{F}_i = \frac{1}{(\delta_i^2 - e_i^2)^2} \Big[ (\delta_i \mathcal{F}_i - e_i \ddot{\delta}_i)(\delta_i^2 - e_i^2) \\
- (\delta_i \dot{e}_i - e_i \dot{\delta}_i)(2\delta_i \dot{\delta}_i - 2e_i \dot{e}_i) \Big] \label{eq:F_tilde_full}
\end{multline}
\begin{align}
\tilde{G}_i &= \frac{\delta_i(t)}{\delta_i(t)^2 - e_i(t)^2} \label{eq:G_tilde_full} \\
\tilde{D}_i &= \frac{\delta_i(t)\mathcal{D}_i}{\delta_i(t)^2 - e_i(t)^2} = \frac{\delta_i(t)D_i(t)}{\delta_i(t)^2 - e_i(t)^2} \label{eq:D_tilde_full}
\end{align}
Note that since the Prescribed Performance Control (PPC) methodology guarantees $\delta_i(t)^2 - e_i(t)^2 \neq 0$, the control gain $\tilde{G}_i$ is always non-zero and well-defined.

\subsubsection{Sliding Surface Design}
For the transformed system described in \eqref{eq:transformed_dynamics_full}, a sliding surface $S_i(t)$ is designed as follows:
\begin{equation}
S_i(t) = \dot{\xi}_i(t) + h_i(\xi_i(t))
\label{eq:sliding_surface_full}
\end{equation}
where $h_i(\xi_i)$ is a specially designed nonlinear function given by:
\begin{equation}
h_i(\xi_i(t)) = k_{1,i} \left( 1 + \alpha_{1,i} |\xi_i(t)|^{1-\gamma_{1,i}} \right)^2 \xi_i(t)^{\gamma_{1,i}}
\label{eq:h_function_full}
\end{equation}
Here, $k_{1,i} > 0$, $\alpha_{1,i} > 0$, and $0 < \gamma_{1,i} < 1$ are design parameters. The exponent $\gamma_{1,i} = p_{i,1}/q_{i,1}$ satisfies the conditions $0 < p_{i,1} < q_{i,1}$, where $q_{i,1}$ is a positive odd integer.

\subsubsection{Control Law Design}
The time derivative of the sliding surface $S_i(t)$ is:
\begin{equation}
\dot{S}_i(t) = \ddot{\xi}_i(t) + \frac{\partial h_i(\xi_i(t))}{\partial \xi_i(t)} \dot{\xi}_i(t)
\label{eq:S_dot_def_full}
\end{equation}
Substituting the transformed dynamics from \eqref{eq:transformed_dynamics_full} into \eqref{eq:S_dot_def_full} yields:
\begin{equation}
\dot{S}_i(t) = \tilde{F}_i + \tilde{G}_i U_{i}(t) + \tilde{D}_i + \frac{\partial h_i(\xi_i(t))}{\partial \xi_i(t)} \dot{\xi}_i(t)
\label{eq:S_dot_expanded_full}
\end{equation}
To ensure the sliding variable $S_i(t)$ converges to zero rapidly and maintains strong robustness, the following nonlinear reaching law is employed:
\begin{equation}
\dot{S}_i(t) = -k_{2,i} \left( 1 + \alpha_{2,i} |S_i(t)|^{1-\gamma_{2,i}} \right)^2 S_i(t)^{\gamma_{2,i}} - \eta_i \text{sign}(S_i(t))
\label{eq:reaching_law_full}
\end{equation}
where $k_{2,i} > 0$, $\alpha_{2,i} > 0$, $0 < \gamma_{2,i} < 1$, and $\eta_i > 0$ are design parameters. By equating \eqref{eq:S_dot_expanded_full} and \eqref{eq:reaching_law_full}, we can solve for the control input $U_{i}(t)$:
\begin{multline}
\tilde{F}_i + \tilde{G}_i U_{i}(t) + \tilde{D}_i + \frac{\partial h_i}{\partial \xi_i} \dot{\xi}_i = \\ -k_{2,i} \left( 1 + \alpha_{2,i} |S_i|^{1-\gamma_{2,i}} \right)^2 S_i^{\gamma_{2,i}} - \eta_i \text{sign}(S_i)
\end{multline}
Considering that the robust term $\eta_i \text{sign}(S_i)$ is designed to counteract the lumped disturbance $\tilde{D}_i$, the control law is formulated as:
\begin{equation}
\begin{split}
U_{i}(t) = \frac{1}{\tilde{G}_i} \bigg[ &-\tilde{F}_i - \frac{\partial h_i(\xi_i)}{\partial \xi_i} \dot{\xi}_i \\
& - k_{2,i} \left( 1 + \alpha_{2,i} |S_i|^{1-\gamma_{2,i}} \right)^2 S_i^{\gamma_{2,i}} \\
& - \eta_i \text{sign}(S_i) \bigg]
\end{split}
\label{eq:control_law_intermediate_full}
\end{equation}
Substituting the expression for $\tilde{G}_i$ from \eqref{eq:G_tilde_full} back into the equation, the explicit form of the control law is obtained as:
\begin{equation}
\begin{split}
U_{i}(t) = \frac{\delta_i^2 - e_i^2}{\delta_i} \bigg[ &-\tilde{F}_i - \frac{\partial h_i(\xi_i)}{\partial \xi_i} \dot{\xi}_i \\
& - k_{2,i} \left( 1 + \alpha_{2,i} |S_i|^{1-\gamma_{2,i}} \right)^2 S_i^{\gamma_{2,i}} \\
& - \eta_i \text{sign}(S_i) \bigg]
\end{split}
\label{eq:control_law_final_full}
\end{equation}

\subsubsection{Fixed-Time Convergence Analysis}
\paragraph{Convergence Analysis of the Sliding Surface \texorpdfstring{$S_i(t)$}{Si(t)}}
To analyze the reaching phase, a Lyapunov function candidate is selected as:
\begin{equation}
V_S(S_i) = \frac{1}{2} S_i(t)^2
\label{eq:lyap_S_full}
\end{equation}
The time derivative is $\dot{V}_S(S_i) = S_i(t) \dot{S}_i(t)$. Substituting the actual control law from \eqref{eq:control_law_intermediate_full} into the sliding surface dynamics \eqref{eq:S_dot_expanded_full} yields:
\begin{align}
\dot{S}_i(t) &= \tilde{F}_i + \tilde{G}_i \left( \frac{1}{\tilde{G}_i} \left[ -\tilde{F}_i - \frac{\partial h_i}{\partial \xi_i} \dot{\xi}_i - k_{2,i} \left( \dots \right)^2 S_i^{\gamma_{2,i}} - \eta_i \text{sign}(S_i) \right] \right) \nonumber \\ 
& \quad + \tilde{D}_i + \frac{\partial h_i}{\partial \xi_i} \dot{\xi}_i \nonumber \\
&= \tilde{D}_i - k_{2,i} \left( 1 + \alpha_{2,i} |S_i|^{1-\gamma_{2,i}} \right)^2 S_i^{\gamma_{2,i}} - \eta_i \text{sign}(S_i)
\label{eq:S_dot_simplified_full}
\end{align}
Let $\Delta_S(t) = \tilde{D}_i$ represent the lumped uncertainty, which is assumed to be bounded by $|\Delta_S(t)| \leq \Delta_{S,max}$. Then, the derivative of the Lyapunov function becomes:
\begin{equation}
\dot{V}_S(S_i) = S_i(t) \Delta_S(t) - k_{2,i} \left( 1 + \alpha_{2,i} |S_i|^{1-\gamma_{2,i}} \right)^2 |S_i|^{\gamma_{2,i}+1} - \eta_i |S_i(t)|
\label{eq:V_dot_S_full_full}
\end{equation}
The robust gain $\eta_i$ is chosen to satisfy $\eta_i \geq \Delta_{S,max}$. Consequently, $S_i(t) \Delta_S(t) - \eta_i |S_i(t)| \leq (\Delta_{S,max} - \eta_i) |S_i(t)| \leq 0$. The derivative can be upper-bounded as:
\begin{equation}
\dot{V}_S(S_i) \leq -k_{2,i} \left( 1 + \alpha_{2,i} |S_i(t)|^{1-\gamma_{2,i}} \right)^2 |S_i(t)|^{\gamma_{2,i}+1}
\end{equation}
Substituting $|S_i| = \sqrt{2V_S}$ and using the inequality $(1+x)^2 \geq 1+x^2$ for $x \ge 0$, we get:
\begin{align}
\dot{V}_S &\leq -k_{2,i} \left[ 1 + \alpha_{2,i}^2 \left( (2V_S)^{\frac{1-\gamma_{2,i}}{2}} \right)^2 \right] (2V_S)^{\frac{\gamma_{2,i}+1}{2}} \nonumber \\
&\leq -k_{2,i} (2V_S)^{\frac{\gamma_{2,i}+1}{2}} - k_{2,i} \alpha_{2,i}^2 (2V_S)^{1-\gamma_{2,i}} (2V_S)^{\frac{\gamma_{2,i}+1}{2}} \nonumber \\
&\leq -k_{2,i} 2^{\frac{\gamma_{2,i}+1}{2}} V_S^{\frac{\gamma_{2,i}+1}{2}} - k_{2,i} \alpha_{2,i}^2 2^{\frac{3-\gamma_{2,i}}{2}} V_S^{\frac{3-\gamma_{2,i}}{2}}
\label{eq:lyap_deriv_S_final_full}
\end{align}
According to Definition~\ref{def_fixed_time_stability}, the convergence time for the reaching phase, $T_{S,max}$, is upper-bounded by:
\begin{equation}
T_{S,max} \leq \frac{2^{\frac{1-\gamma_{2,i}}{2}}}{k_{2,i}(1-\gamma_{2,i})} \left( 1 + \frac{1}{\alpha_{2,i}^2 2^{1-\gamma_{2,i}}} \right)
\label{eq:time_S_full}
\end{equation}

\paragraph{Convergence Analysis of the Transformed Error \texorpdfstring{$\xi_i(t)$}{nu_check(t)}}
Once $S_i(t) \equiv 0$, the sliding mode dynamics are $\dot{\xi}_i(t) = -h_i(\xi_i(t))$. We select a Lyapunov function $V_{\xi} = \frac{1}{2} \xi_i(t)^2$. Its time derivative is:
\begin{equation}
\dot{V}_{\check{\nu}}(\xi_i) = -k_{1,i} \left( 1 + \alpha_{1,i} |\xi_i|^{1-\gamma_{1,i}} \right)^2 |\xi_i|^{\gamma_{1,i}+1}
\end{equation}
Following a similar derivation, we arrive at:
\begin{equation}
\dot{V}_{\xi} \leq -k_{1,i} 2^{\frac{\gamma_{1,i}+1}{2}} V_{\xi}^{\frac{\gamma_{1,i}+1}{2}} - k_{1,i} \alpha_{1,i}^2 2^{\frac{3-\gamma_{1,i}}{2}} V_{\xi}^{\frac{3-\gamma_{1,i}}{2}}
\label{eq:lyap_deriv_nu_final_full}
\end{equation}
The convergence time $T_{\xi,max}$ is thus upper-bounded by:
\begin{equation}
T_{\xi,max} \leq \frac{2^{\frac{1-\gamma_{1,i}}{2}}}{k_{1,i}(1-\gamma_{1,i})} \left( 1 + \frac{1}{\alpha_{1,i}^2 2^{1-\gamma_{1,i}}} \right)
\label{eq:time_nu_full}
\end{equation}

\paragraph{Total Convergence Time of the System}
The total time for the transformed error $\xi_i(t)$ to converge to zero is bounded by $T_{total,max}$:
\begin{equation}
T_{total,max} = T_{S,max} + T_{\xi,max}
\label{eq:time_total_full}
\end{equation}
For $t \geq T_{total,max}$, we have $\xi_i(t) \equiv 0$, which implies $e_i(t) = 0$. The proposed FTSMC strategy therefore guarantees fixed-time convergence of the tracking error.

\subsection{Tracking Error-Based Fixed-Time Composite Disturbance Observer Design}
To facilitate high-precision and rapid estimation of the unknown composite disturbances inherent in the system, this paper proposes a Tracking Error-Based Fixed-Time Composite Disturbance Observer (TFxTCDO). Unlike traditional observers that rely on full-state measurements or precise system modeling, TFxTCDO leverages tracking error information to directly associate disturbance estimation with control performance. Specifically, an tracking error variable $v_{e,i} = \dot{\nu}_i - \dot{v}_{ref,i}$, which characterizes the deviation between the actual state derivative and the output of a reference model, the TFxTCDO is designed to track both this variable ($z_{1,i} = v_{e,i}$) and its derivative ($z_{2,i} = \dot{v}_{e,i}$) with fixed-time precision. This architecture enables the accurate reconstruction and estimation of the disturbance $D_i(t)$.

Consider the original system dynamics $\ddot{\nu}_i = U_i + F_i + D_i$. We introduce a reference estimator output, denoted by $\dot{v}_{ref,i}$, to construct the disturbance observation error:
\begin{equation}
v_{e,i} = \dot{\nu}_i - \dot{v}_{ref,i}
\end{equation}
The reference estimator dynamics are designed as:
\begin{equation}
\ddot{v}_{ref,i} = U_i + F_i + \frac{v_{e,i}}{\epsilon_i}
\end{equation}
where $\epsilon_i$ is a small positive design parameter. From the equations above, the dynamics of $v_{e,i}$ can be derived as:
\begin{align}
\dot{v}_{e,i} &= \ddot{\nu}_i - \ddot{v}_{ref,i} \nonumber \\
&= (U_i + F_i + D_i) - \left(U_i + F_i + \frac{v_{e,i}}{\epsilon_i}\right) \nonumber \\
&= D_i - \frac{v_{e,i}}{\epsilon_i}
\end{align}
This result reveals that $v_{e,i}$ serves as a crucial intermediate variable intrinsically related to the disturbance $D_i$. If both $v_{e,i}$ and its derivative $\dot{v}_{e,i}$ can be accurately estimated, the disturbance $D_i$ can be reconstructed as follows:
\begin{equation}
D_i = \dot{v}_{e,i} + \frac{v_{e,i}}{\epsilon_i}
\label{eq:disturbance_reconstruction_full}
\end{equation}
To this end, a new state vector is defined based on $v_{e,i}$:
\begin{align}
z_{1,i} &= v_{e,i} \\
z_{2,i} &= \dot{v}_{e,i} = D_i - \frac{v_{e,i}}{\epsilon_i} = D_i - \frac{z_{1,i}}{\epsilon_i}
\end{align}
The dynamics of this new state can be expressed as a second-order system:
\begin{equation}
\begin{cases}
    \dot{z}_{1,i} = z_{2,i} \\
    \dot{z}_{2,i} = \dot{D}_i - \frac{\dot{z}_{1,i}}{\epsilon_i} = \dot{D}_i - \frac{z_{2,i}}{\epsilon_i}  \label{eq:err_dyn_model}
\end{cases}
\end{equation}
Let $\hat{z}_{1,i}$ and $\hat{z}_{2,i}$ be the estimates of $z_{1,i}$ and $z_{2,i}$, respectively. The estimation errors are defined as $\tilde{z}_{1,i} = z_{1,i} - \hat{z}_{1,i}$ and $\tilde{z}_{2,i} = z_{2,i} - \hat{z}_{2,i}$. The proposed FxTCDO is designed with the following dynamics, where we use the notation $\text{sign}(x)|x|^y$ for clarity:
\begin{align}
\dot{\hat{z}}_{1,i} = \hat{z}_{2,i} &+ \lambda_{1,i} \text{sign}(\tilde{z}_{1,i})|\tilde{z}_{1,i}|^{\alpha_{1,i}} + \lambda_{2,i} \text{sign}(\tilde{z}_{1,i})|\tilde{z}_{1,i}|^{\alpha_{2,i}} \nonumber \\ 
& + \lambda_{3,i} \tilde{z}_{1,i} \label{eq:obs_dyn_1} \\
\dot{\hat{z}}_{2,i} = -\frac{\hat{z}_{2,i}}{\epsilon_i} &+ \mu_{1,i} \text{sign}(\tilde{z}_{2,i})|\tilde{z}_{2,i}|^{\beta_{1,i}} + \mu_{2,i} \text{sign}(\tilde{z}_{2,i})|\tilde{z}_{2,i}|^{\beta_{2,i}} \nonumber \\
& + \mu_{3,i} \tilde{z}_{2,i} + K_{s,i} \text{sign}(\tilde{z}_{2,i}) \label{eq:obs_dyn_2}
\end{align}

The parameters $\lambda_{j,i} > 0$, $\mu_{j,i} > 0$, and $K_{s,i} > 0$ are the observer gains. The exponents satisfy $0 < \alpha_{1,i} < 1$, $\alpha_{2,i} > 1$, $0 < \beta_{1,i} < 1$, and $\beta_{2,i} > 1$.

Consequently, the dynamics of the estimation error are derived as:
\begin{align}
\dot{\tilde{z}}_{1,i} = \tilde{z}_{2,i} &- \big( \lambda_{1,i} \text{sign}(\tilde{z}_{1,i})|\tilde{z}_{1,i}|^{\alpha_{1,i}} + \lambda_{2,i} \text{sign}(\tilde{z}_{1,i})|\tilde{z}_{1,i}|^{\alpha_{2,i}} \nonumber \\ 
& + \lambda_{3,i} \tilde{z}_{1,i} \big) \label{eq:z1_tilde_dot_full} \\
\dot{\tilde{z}}_{2,i} = \dot{D}_i - \frac{\tilde{z}_{2,i}}{\epsilon_i} &- \big( \mu_{1,i} \text{sign}(\tilde{z}_{2,i})|\tilde{z}_{2,i}|^{\beta_{1,i}} + \mu_{2,i} \text{sign}(\tilde{z}_{2,i})|\tilde{z}_{2,i}|^{\beta_{2,i}} \nonumber \\
& + \mu_{3,i} \tilde{z}_{2,i} + K_{s,i} \text{sign}(\tilde{z}_{2,i}) \big) \label{eq:z2_tilde_dot_full}
\end{align}

\subsubsection{Stability Proof of the Observer}
The stability of the proposed FxTCDO is established through a two-step analysis based on the cascade structure of the estimation error dynamics.

\paragraph{Step 1: Analysis of the \texorpdfstring{$\tilde{z}_{2,i}$}{z2_tilde} Subsystem}
Consider the Lyapunov function candidate $V_{2,i} = \frac{1}{2} \tilde{z}_{2,i}^2$. Its time derivative along the trajectory of \eqref{eq:z2_tilde_dot_full} is:
\begin{equation}
\begin{split}
    \dot{V}_{2,i} ={}& \tilde{z}_{2,i} \dot{D}_i - \frac{\tilde{z}_{2,i}^2}{\epsilon_i} - \mu_{1,i} |\tilde{z}_{2,i}|^{\beta_{1,i}+1} \\
    & - \mu_{2,i} |\tilde{z}_{2,i}|^{\beta_{2,i}+1} - \mu_{3,i} \tilde{z}_{2,i}^2 - K_{s,i} |\tilde{z}_{2,i}|
\end{split}
\end{equation}

Assuming that the derivative of the disturbance is bounded, $|\dot{D}_i| \leq L_D > 0$, and selecting the gain $K_{s,i} > L_D$, the derivative of the Lyapunov function can be upper-bounded as:
\begin{multline}
\dot{V}_{2,i} \leq -(K_{s,i} - L_D)|\tilde{z}_{2,i}| - \left(\frac{1}{\epsilon_i} + \mu_{3,i}\right)\tilde{z}_{2,i}^2 \\
- \mu_{1,i}|\tilde{z}_{2,i}|^{\beta_{1,i}+1} - \mu_{2,i}|\tilde{z}_{2,i}|^{\beta_{2,i}+1}
\end{multline}
For fixed-time stability analysis, we focus on the dominant terms:
\begin{equation}
\dot{V}_{2,i} \leq -c_{1,i} V_{2,i}^{\frac{\beta_{1,i}+1}{2}} - c_{2,i} V_{2,i}^{\frac{\beta_{2,i}+1}{2}}
\label{eq:V2_dot_final_full}
\end{equation}
where $c_{1,i} = 2^{\frac{\beta_{1,i}+1}{2}}\mu_{1,i}$ and $c_{2,i} = 2^{\frac{\beta_{2,i}+1}{2}}\mu_{2,i}$. Since the exponents satisfy $\frac{1}{2} < \frac{\beta_{1,i}+1}{2} < 1$ and $\frac{\beta_{2,i}+1}{2} > 1$, the inequality \eqref{eq:V2_dot_final_full} ensures that $\tilde{z}_{2,i}$ converges to zero in a fixed time.

\paragraph{Step 2: Stability Analysis of the \texorpdfstring{$\tilde{z}_{1,i}$}{z1_tilde} Subsystem}
Once the state $\tilde{z}_{2,i}$ converges to zero, as guaranteed in a fixed time by the analysis in Step 1, the dynamics of the error state $\tilde{z}_{1,i}$ are governed by the following simplified equation:
\begin{equation}
\dot{\tilde{z}}_{1,i} = -(\lambda_{1,i} \lceil \tilde{z}_{1,i} \rceil^{\alpha_{1,i}} + \lambda_{2,i} \lceil \tilde{z}_{1,i} \rceil^{\alpha_{2,i}} + \lambda_{3,i} \tilde{z}_{1,i})
\label{eq:z1_tilde_dot_simplified}
\end{equation}
where the notation $\lceil x \rceil^{\alpha} \triangleq |x|^{\alpha}\text{sgn}(x)$ is used. To analyze the stability of this subsystem, we consider the Lyapunov function candidate $V_{1,i} = \frac{1}{2}\tilde{z}_{1,i}^2$. The time derivative of $V_{1,i}$ along the trajectories of \eqref{eq:z1_tilde_dot_simplified} is given by:
\begin{equation}
\begin{split}
    \dot{V}_{1,i} &= \tilde{z}_{1,i}\dot{\tilde{z}}_{1,i} \\
    &= -\lambda_{1,i} |\tilde{z}_{1,i}|^{\alpha_{1,i}+1} - \lambda_{2,i} |\tilde{z}_{1,i}|^{\alpha_{2,i}+1} - \lambda_{3,i}\tilde{z}_{1,i}^2 \\
    &= -c_{3,i}V_{1,i}^{\frac{\alpha_{1,i}+1}{2}} - c_{4,i}V_{1,i}^{\frac{\alpha_{2,i}+1}{2}} - 2\lambda_{3,i}V_{1,i}
\end{split}
\label{eq:V1_dot_full}
\end{equation}
where $c_{3,i} = 2^{\frac{\alpha_{1,i}+1}{2}}\lambda_{1,i}$ and $c_{4,i} = 2^{\frac{\alpha_{2,i}+1}{2}}\lambda_{2,i}$. Since the term $-2\lambda_{3,i}V_{1,i}$ is non-positive (given $\lambda_{3,i}>0$), the following inequality holds:
\begin{equation}
\dot{V}_{1,i} \le -c_{3,i}V_{1,i}^{\frac{\alpha_{1,i}+1}{2}} - c_{4,i}V_{1,i}^{\frac{\alpha_{2,i}+1}{2}}
\label{eq:V1_dot_final}
\end{equation}

This inequality is in the standard form for fixed-time stability. The parameter constraints $0 < \alpha_{1,i} < 1$ and $\alpha_{2,i} > 1$ ensure that the exponents satisfy the conditions for fixed-time convergence, namely $\frac{1}{2} < \frac{\alpha_{1,i}+1}{2} < 1$ and $\frac{\alpha_{2,i}+1}{2} > 1$. Consequently, based on the fixed-time stability lemma, the state $\tilde{z}_{1,i}$ is also guaranteed to converge to zero in a fixed time.

\subsection*{Overall Disturbance Estimation Error Convergence}
Having established the fixed-time convergence of both internal observer states, $\tilde{z}_{1,i}$ and $\tilde{z}_{2,i}$, we can now analyze the overall disturbance estimation error. The disturbance estimate $\hat{D}_i$ is reconstructed as:
\begin{equation}
\hat{D}_i = \hat{z}_{2,i} + \frac{\hat{z}_{1,i}}{\epsilon_i}
\end{equation}
Therefore, the disturbance estimation error, $\tilde{D}_i = D_i - \hat{D}_i$, is given by:
\begin{equation}
\tilde{D}_i = \tilde{z}_{2,i} + \frac{\tilde{z}_{1,i}}{\epsilon_i}
\end{equation}
Since both $\tilde{z}_{1,i}$ and $\tilde{z}_{2,i}$ converge to zero in a fixed time, it follows directly that the overall disturbance estimation error, $\tilde{D}_i$, will also converge to zero within a fixed time. This completes the proof of the observer's fixed-time convergence property.

\begin{remark}
Conventional Prescribed Performance Control (PPC) constrains the error dynamics within a predefined performance envelope, but typically relies on asymptotic or exponential convergence. Conversely, while Sliding Mode Control (SMC) is renowned for its robustness, it often lacks a systematic mechanism for enforcing performance constraints. This work synergizes the error-constraining capability of PPC with the fixed-time convergence property of SMC. By introducing a transformed error (\ref{eq:transformed_error_full}) and a nonlinear sliding surface (\ref{eq:sliding_surface_full}), our approach simultaneously achieves prescribed performance and fixed-time convergence within a unified framework.
\end{remark}

\begin{remark}
This paper achieves fixed-time convergence by meticulously designing the nonlinear sliding surface and the reaching law (\ref{eq:reaching_law_full}). Specifically, the combination of fractional power terms (e.g., $|S_i|^{\gamma_{2,i}}$) and higher-order nonlinear terms (i.e., $(1+\alpha_{2,i}|S_i|^{1-\gamma_{2,i}})^2$) plays a crucial role. This structure not only accelerates the convergence process but also provides additional degrees of freedom through parameters $\alpha_{2,i}$ and $k_{2,i}$ to optimize the trade-off between convergence time and control input smoothness.
\end{remark}

\begin{remark}
The proposed control law (\ref{eq:control_law_final_full}) incorporates a state-dependent gain structure, overcoming the limitation of conventional SMC which often employs a fixed switching gain. The adaptive nature of this gain ensures that when the system state is far from the sliding surface ($|S_i|$ is large), the control effort is automatically intensified to expedite convergence. Conversely, as the state approaches the sliding surface ($|S_i|$ is small), the gain is smoothly reduced, which contributes to chattering alleviation.
\end{remark}

\begin{remark}
A key advantage of the control architecture is its independence from an explicit system model formulation. The controller's design and implementation do not require prior knowledge of the system's dynamic parameters or the disturbance's specific structure. Instead, it is driven entirely by real-time observation errors, which endows it with strong model-free characteristics and enhances its applicability across a diverse range of systems.
\end{remark}

\begin{remark}
The proposed TFxTCDO, as formulated in \eqref{eq:obs_dyn_1}--\eqref{eq:obs_dyn_2}, introduces an innovative departure from traditional observer design philosophies. A key characteristic of this observer is that its design \textit{does not depend on the complex, nonlinear dynamic model of the quadrotor presented in \eqref{eq:trans_x}--\eqref{eq:rot_psi}}, but rather on a \textit{simplified, derived error dynamic model given by \eqref{eq:err_dyn_model}}. 
This method departs from traditional observer designs by leveraging tracking error to construct the estimation law, thereby tightly coupling disturbance estimation with control objectives and enhancing system responsiveness. 
By using directly measurable tracking error signals that inherently reflect system perturbations, it eliminates the need for detailed knowledge of the plant's parameters (e.g., mass, inertia) and nonlinearities, thus demonstrating strong robustness against model uncertainties. 
Moreover, the observer ensures fixed-time convergence, enabling rapid and bounded estimation of composite disturbances irrespective of initial conditions.
\end{remark}
\section{Simulation Results}
\label{sec:simulations}

To validate the efficacy and superiority of the proposed control strategy, a series of numerical simulations were conducted. The quadrotor dynamics are based on the mathematical model detailed in the preceding sections. To ensure the reproducibility of the results, the system and control parameters used throughout the simulations are explicitly listed in Table~\ref{tab:physical_params} and Table~\ref{tab:control_params_combined}, respectively.

%

\begin{table}[htbp]
\centering
\caption{Physical Model Parameters.}
\label{tab:physical_params}
\begin{tabular}{l l l}
\toprule
Symbol & Value & Units \\
\midrule
$m$ & 0.8 & kg \\
$g$ & 9.81 & m/s$^2$ \\
$I_{xx}$ & 0.082 & kg$\cdot$m$^2$ \\
$I_{yy}$ & 0.082 & kg$\cdot$m$^2$ \\
$I_{zz}$ & 0.149 & kg$\cdot$m$^2$ \\
$d$ & 0.45 & m \\ 
$b$ & $2 \times 10^{-6}$ & N/(rad/s)$^2$ \\
$d_{\text{drag}}$ & $2.1 \times 10^{-7}$ & N$\cdot$m/(rad/s)$^2$ \\ 
\bottomrule
\end{tabular}
\end{table}


\begin{table}[htbp]
\centering
\caption{Controller Parameters for Position ($i=1,2,3$) and Attitude ($i=4,5,6$) States.}
\label{tab:control_params_combined}
\begin{tabularx}{\linewidth}{l C C} 
\toprule
\textbf{Symbols} & \textbf{Values for Position ($i=1,2,3$)} & \textbf{Values for Attitude ($i=4,5,6$)} \\
\midrule
$\delta_{i0}, \delta_{iT}, l_i, T_i$ & $30, 0.05, 1.15, 8$  & $3, 0.01, 1.25, 5$ \\
\addlinespace 
$k_{1,i}, \gamma_{1,i}, \alpha_{1,i}$ & $5, 7/9, 1/2$ & $40, 7/9, 0.5$ \\
$k_{2,i}, \alpha_{2,i}, \gamma_{2,i}, \eta_i$ & $5, 0.1, 7/9, 1/2$ & $40, 0.5, 7/9, 35$ \\
\addlinespace
$\epsilon_i, \mu_{1,i}, \mu_{2,i}, \mu_{3,i}$ & $0.15, 0.15, 0.15, 0.15$ &0.34, 0.24, 0.24, 0.24 \\ 
\addlinespace
$\beta_{1,i}, \beta_{2,i}, \beta_{3,i}$ & $0.75, 1.2, 1.0$ & $0.8, 1.1, 1.0$ \\
\bottomrule
\end{tabularx}
\end{table}

\subsection{Case 1: Trajectory Tracking Performance without Disturbances or Faults}

In this initial case, we evaluate the fundamental tracking performance of the proposed controller under ideal conditions, i.e., free from external disturbances and actuator faults. The objective is to track a typical time-varying trajectory, which is a circular path in the horizontal plane while maintaining a constant altitude and a constant yaw angle. The reference trajectory is mathematically defined as:
\begin{equation}
\begin{cases}
    \nu_1^d(t) = 8 - 8\cos(\frac{t}{4}) \\
    \nu_2^d(t) = 8\sin(\frac{t}{4}) \\
    \nu_3^d(t) = 10 \\
    \nu_6^d(t) = 1
\end{cases}
\end{equation}

\begin{figure}[htbp]
    \centering

    \subfloat[Position tracking in x-axis.]{
        \includegraphics[width=0.47\linewidth]{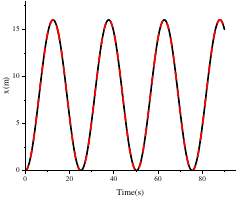}
        \label{fig:case1_x}
    }
    \hfill
    \subfloat[Attitude tracking for $\phi$.]{
        \includegraphics[width=0.47\linewidth]{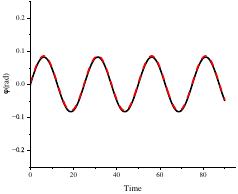}
        \label{fig:case1_phi}
    }

    \vfill 

    \subfloat[Position tracking in y-axis.]{
        \includegraphics[width=0.47\linewidth]{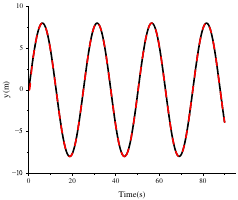}
        \label{fig:case1_y}
    }
    \hfill
    \subfloat[Attitude tracking for $\theta$.]{
        \includegraphics[width=0.47\linewidth]{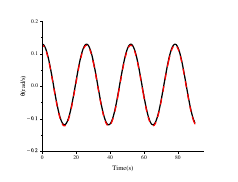}
        \label{fig:case1_theta}
    }
    
    \vfill 

    \subfloat[Position tracking in z-axis.]{
        \includegraphics[width=0.47\linewidth]{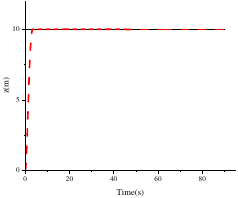}
        \label{fig:case1_z}
    }
    \hfill
    \subfloat[Attitude tracking for $\psi$.]{
        \includegraphics[width=0.47\linewidth]{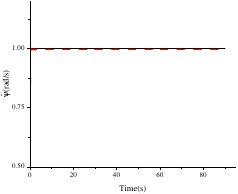}
        \label{fig:case1_psi}
    }

    \caption{Tracking performance of the proposed controller in Case 1. The actual trajectory (red dashed line) is shown to closely follow the desired reference (black solid line). The left column (a, c, e) shows the position tracking, while the right column (b, d, f) shows the attitude tracking.}
    \label{fig:case1_tracking_performance}
\end{figure}

\begin{figure}[htbp]
    \centering

    \subfloat[Linear velocity in x-axis ($\dot{x}$).]{
        \includegraphics[width=0.47\linewidth]{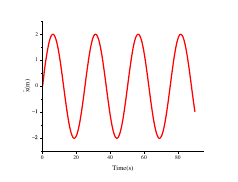}
        \label{fig:case1_x_dot}
    }
    \hfill
    \subfloat[Angular velocity $\dot{\phi}$.]{
        \includegraphics[width=0.47\linewidth]{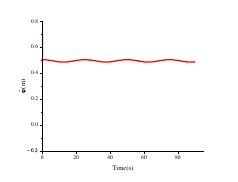}
        \label{fig:case1_phi_dot}
    }

    \vfill 

    \subfloat[Linear velocity in y-axis ($\dot{y}$).]{
        \includegraphics[width=0.47\linewidth]{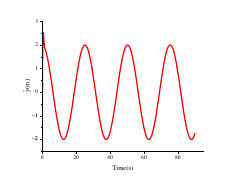}
        \label{fig:case1_y_dot}
    }
    \hfill
    \subfloat[Angular velocity $\dot{\theta}$.]{
        \includegraphics[width=0.47\linewidth]{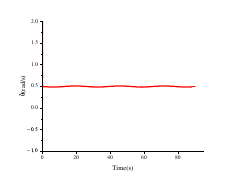}
        \label{fig:case1_theta_dot}
    }
    
    \vfill 

    \subfloat[Linear velocity in z-axis ($\dot{z}$).]{
        \includegraphics[width=0.47\linewidth]{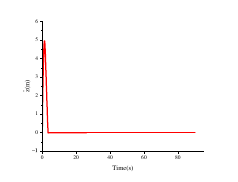}
        \label{fig:case1_z_dot}
    }
    \hfill
    \subfloat[Angular velocity $\dot{\psi}$.]{
        \includegraphics[width=0.47\linewidth]{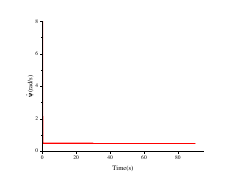}
        \label{fig:case1_psi_dot}
    }

    \caption{Response of the state derivatives in Case 1. The left column (a, c, e) shows the linear velocities, and the right column (b, d, f) shows the angular velocities. All responses are smooth and stable, indicating excellent transient regulation performance.}
    \label{fig:case1_derivatives}
\end{figure}

\begin{figure}[htbp]
    \centering

    \subfloat[Position error in x-axis.]{
        \includegraphics[width=0.47\linewidth]{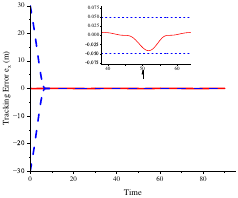}
        \label{fig:case1_error_x}
    }
    \hfill
    \subfloat[Attitude error for $\phi$.]{
        \includegraphics[width=0.47\linewidth]{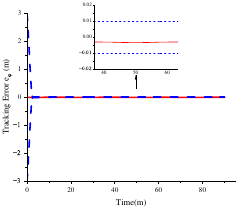}
        \label{fig:case1_error_phi}
    }

    \vfill 

    \subfloat[Position error in y-axis.]{
        \includegraphics[width=0.47\linewidth]{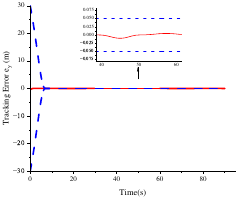}
        \label{fig:case1_error_y}
    }
    \hfill
    \subfloat[Attitude error for $\theta$.]{
        \includegraphics[width=0.47\linewidth]{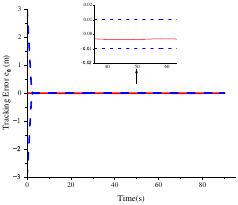}
        \label{fig:case1_error_theta}
    }
    
    \vfill 

    \subfloat[Position error in z-axis.]{
        \includegraphics[width=0.47\linewidth]{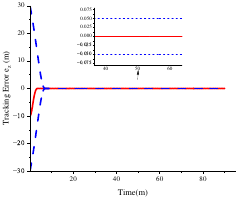}
        \label{fig:case1_error_z}
    }
    \hfill
    \subfloat[Attitude error for $\psi$.]{
        \includegraphics[width=0.47\linewidth]{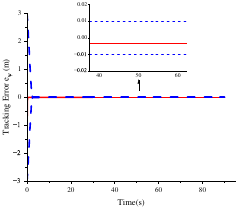}
        \label{fig:case1_error_psi}
    }

    \caption{Evolution of the tracking errors in Case 1. The tracking error (red solid line) for each state is shown to converge rapidly and remain within the prescribed performance boundaries (blue dashed lines), validating the controller's high precision.}
    \label{fig:case1_tracking_errors}
\end{figure}

To provide a comprehensive assessment of the controller's tracking capabilities in this nominal scenario, the simulation results are presented in Figures~\ref{fig:case1_tracking_performance}–\ref{fig:case1_tracking_errors}. These figures respectively illustrate the trajectory tracking of the six position and attitude states ($\nu_1(t)$ to $\nu_6(t)$), the response of their first-order derivatives ($\dot{\nu}_1(t)$ to $\dot{\nu}_6(t)$), and the evolution of the corresponding tracking errors ($e_x(t)$, $e_y(t)$, $e_z(t)$, $e_\varphi(t)$, $e_\theta(t)$, $e_\psi(t)$).

As depicted in Fig.~\ref{fig:case1_tracking_performance}, the proposed controller exhibits outstanding tracking accuracy for all six states. The actual trajectory (red dashed line) almost perfectly overlaps with the desired reference (black solid line), demonstrating excellent phase synchronization and amplitude consistency, particularly for the circular motion. This highlights the control system's adaptability to nonlinear reference trajectories.

Figure~\ref{fig:case1_derivatives} illustrates that the derivatives of all states exhibit smooth and stable responses, devoid of significant overshoot or oscillations. This indicates the controller's excellent transient regulation performance, even under conditions of rapid dynamic changes.

The evolution of the six state tracking errors is presented in Fig.~\ref{fig:case1_tracking_errors}. It can be observed that all error variables not only converge rapidly to a small vicinity of zero but are also \textit{strictly maintained within their predefined performance boundaries (blue dashed lines) at all times}, which validates the effectiveness of the AT-PPF component. Subsequently, the errors are kept within this minimal range, with steady-state errors being approximately zero. This further validates the controller's high tracking precision and robust dynamic response capabilities in the absence of external disturbances.

In summary, the simulation results from Case~1 conclusively demonstrate that the proposed prescribed-time sliding mode fault-tolerant control strategy possesses excellent tracking precision, rapid convergence speed, and strong stability characteristics under ideal conditions. This establishes a reliable baseline for evaluating the controller's performance in the subsequent, more challenging scenarios involving disturbances and faults.

\subsection{Case 2: Fault-Tolerant Control Performance under Actuator Faults and Disturbances}

To rigorously evaluate the fault-tolerant capabilities and robustness of the proposed control strategy in the presence of external disturbances and actuator faults, a second case (Case 2) was designed. Building upon the same reference trajectory and initial conditions as in Case 1, we introduce simultaneous adverse conditions at $t=40$~s. Specifically, multiplicative faults are applied to the four motors with efficiency factors of 0.85, 0.80, 0.75, and 0.70, respectively, simulating a scenario of significant actuator performance degradation. Concurrently, the system is subjected to a composite disturbance signal, defined as $d(t) = [1 + e^{-t/2} + \sin(t), \dots]^T$, which incorporates both low-frequency and high-frequency components to represent a typical uncertain disturbance environment.

\begin{figure}[htbp]
    \centering

    \subfloat[Attitude angle $\phi$ response.]{
        \includegraphics[width=0.47\linewidth]{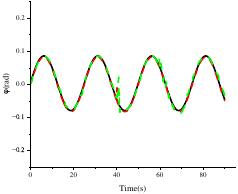} 
        \label{fig:case2_phi}
    }
    \hfill
    \subfloat[Angular velocity $\dot{\phi}$ response.]{
        \includegraphics[width=0.47\linewidth]{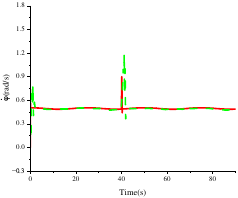} 
        \label{fig:case2_phi_dot}
    }

    \vfill 

    \subfloat[Attitude angle $\theta$ response.]{
        \includegraphics[width=0.47\linewidth]{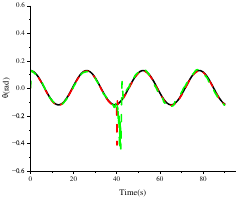} 
        \label{fig:case2_theta}
    }
    \hfill
    \subfloat[Angular velocity $\dot{\theta}$ response.]{
        \includegraphics[width=0.47\linewidth]{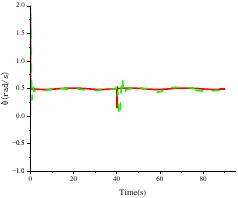} 
        \label{fig:case2_theta_dot}
    }
    
    \vfill 

    \subfloat[Attitude angle $\psi$ response.]{
        \includegraphics[width=0.47\linewidth]{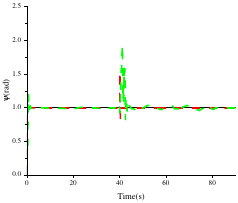} 
        \label{fig:case2_psi}
    }
    \hfill
    \subfloat[Angular velocity $\dot{\psi}$ response.]{
        \includegraphics[width=0.47\linewidth]{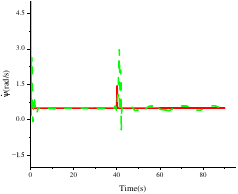} 
        \label{fig:case2_psi_dot}
    }

    \caption{Comparative tracking performance in Case 2 under simultaneous actuator faults and composite disturbances. The legend applies to all subplots. The solid black line represents the desired reference trajectory, the dashed red line shows the response of the proposed method, and the dash-dotted green line corresponds to the SMC with NLO method from \cite{Ref_to_Compared_Method}. The left column (a, c, e) shows the attitude angle responses ($\phi, \theta, \psi$), while the right column (b, d, f) shows the corresponding angular velocity responses.}
    \label{fig:case2_comparison}
\end{figure}

The simulation results, presented in Fig.~\ref{fig:case2_comparison}, reveal the system's dynamic response to these abrupt changes. Around the 40-second mark, all flight states exhibit instantaneous fluctuations. The attitude states ($\nu_4(t)$, $\nu_5(t)$, $\nu_6(t)$) show a brief deviation from their desired trajectories. However, owing to the synergistic action of the controller's rapid compensation mechanism and the integrated disturbance observer, the system adeptly recovers and returns to a stable tracking state within a remarkably short period. Notably, this recovery is achieved without significant oscillations or error accumulation, and the corresponding state derivatives also demonstrate a smooth dynamic restoration process.

\begin{figure}[htbp]
    \centering

    \subfloat[Comparison for the $\phi$-channel.]{
        \includegraphics[width=0.9\linewidth]{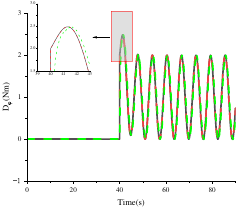} 
        \label{fig:comp_obs_phi}
    }
    \vfill 

    \subfloat[Comparison for the $\theta$-channel.]{
        \includegraphics[width=0.9\linewidth]{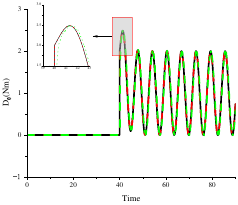}
        \label{fig:comp_obs_theta}
    }
    \vfill

    \subfloat[Comparison for the $\psi$-channel.]{
        \includegraphics[width=0.9\linewidth]{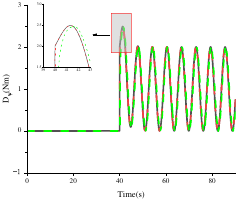}
        \label{fig:comp_obs_psi}
    }

    \caption{Comparative performance of the proposed observer and the observer from \cite{Ref_to_Compared_Method}. In each subplot, the actual disturbance is shown as a dashed line, while the estimations from the proposed method (solid line) and the method from\cite{Ref_to_Compared_Method} (dash-dotted line) are compared.}
    \label{fig:observer_comparison}
\end{figure}

For comparative validation, we implemented the nonlinear observer proposed in \cite{Ref_to_Compared_Method}, which was originally designed for fault-tolerant control of quadrotor UAVs. Although this observer demonstrates a certain level of disturbance estimation capability, its performance under the same adverse conditions is evidently inferior to that of our proposed scheme. As illustrated in the comparative plots in Fig.~\ref{fig:observer_comparison}, the recovery time of the competing method is significantly longer, and the state variables exhibit larger fluctuations. Moreover, its estimation error converges more slowly, and its disturbance reconstruction suffers from noticeable latency. These deficiencies are particularly evident during the initial transient phase following disturbance injection, where the estimation output from \cite{Ref_to_Compared_Method} deviates markedly from the true values.



In stark contrast, our proposed fixed-time disturbance observer (TFxTCDO) demonstrates a markedly superior performance, as illustrated in Fig.~\ref{fig:observer_comparison}. The superiority is most evident in the transient response at the fault-injection instant ($t=40$~s). The proposed TFxTCDO responds almost instantaneously to the abrupt change, with its estimation (solid line) rapidly converging to the actual disturbance value (dashed line) with minimal delay. Conversely, the observer from [34] (dash-dotted line) exhibits a noticeable latency and a significantly slower convergence rate, resulting in a large transient estimation error. Furthermore, in the post-fault phase, our observer maintains a high degree of consistency with the actual disturbance, indicating negligible steady-state error. This accurate and timely estimation provides a solid foundation for the controller's compensation action, which is crucial for the swift attitude recovery shown in Fig.~\ref{fig:case2_comparison}. These findings collectively underscore the superior estimation accuracy, faster convergence, and enhanced robustness offered by the proposed observer design.

\subsection{Case 3: Validation of Fault Tolerance under Parameter Uncertainties}

Having demonstrated the framework's effectiveness against external disturbances and actuator faults, we next assess its \textit{robustness} against internal modeling errors. To this end, a third case (Case 3) was designed. Building upon the scenario of Case 2, this case introduces significant parameter uncertainties of $\pm20\%$ to key physical parameters, such as the aircraft's mass and moments of inertia, to simulate real-world deviations. These perturbed parameters are detailed in Table~\ref{tab:param_uncertainty}. All other fault and disturbance conditions are kept identical to those in Case 2: at $t=40$~s, multiplicative faults (with degradation factors of 0.85, 0.80, 0.75, and 0.70) are applied to the four actuators, concurrent with the injection of the composite disturbance signal $d(t) = [1 + e^{-t/2} + \sin(t), \dots]^T$.

\begin{table}[htbp]
\centering
\caption{Time-Varying Parameter Uncertainties Introduced in Case 3.}
\label{tab:param_uncertainty}
\begin{tabular}{l c l} 
\toprule
\textbf{Parameter} & \textbf{Nominal Value} & \textbf{Actual Value (Time-Varying)} \\
\midrule
$m$ & 0.8 kg & $0.8 \cdot (1 + 0.2\sin(0.1t))$ kg \\
\addlinespace
$I_{xx}$ & 0.082 kg$\cdot$m$^2$ & $0.082 \cdot (1 + 0.2\sin(0.1t))$ kg$\cdot$m$^2$ \\
\addlinespace
$I_{yy}$ & 0.082 kg$\cdot$m$^2$ & $0.082 \cdot (1 + 0.2\sin(0.1t))$ kg$\cdot$m$^2$ \\
\addlinespace
$I_{zz}$ & 0.149 kg$\cdot$m$^2$ & $0.149 \cdot (1 + 0.2\sin(0.1t))$ kg$\cdot$m$^2$ \\
\bottomrule
\end{tabular}
\end{table}

Crucially, the simulation is conducted without any re-tuning of the controller parameters. Despite these compounded challenges, the simulation results demonstrate that the system maintains commendable tracking performance. Although slight fluctuations are observed in the state variables around the disturbance injection point ($t=40$~s), all states re-converge rapidly and sustain stable operation. The overall profiles of the state trajectories and error curves are largely consistent with those presented in Case 2; therefore, for the sake of brevity, they are not repeated here.

To provide a quantitative assessment of the performance, three representative error metrics were calculated: the Mean Squared Error (MSE), the Root Mean Squared Error (RMSE), and the Integral of Squared Error (ISE). Table~\ref{tab:error_metrics} presents a comparative summary of these metrics for the primary position tracking channels. The results indicate that, even in the presence of substantial parameter uncertainties, the performance metrics exhibit only a marginal increase, ranging from approximately 2\% to 12\%. The overall error values remain within a highly acceptable range, robustly validating the characteristic of the proposed control method and confirming its ability to sustain high performance despite modeling inaccuracies.

\begin{table}[htbp]
\centering
\caption{Quantitative Error Metrics for Position and Attitude Tracking.}
\label{tab:error_metrics}
\begin{tabular}{l c c c c} 
\toprule
\textbf{Metric} & \textbf{Channel} & \textbf{Case 2} & \textbf{Case 3} & \textbf{Increase (\%)} \\
\midrule
\multirow{6}{*}{MSE} & $x$ & 0.027 & 0.029 & 7.41 \\
& $y$ & 0.024 & 0.026 & 8.33 \\
& $z$ & 0.016 & 0.018 & 12.50 \\
\addlinespace
& $\varphi$ & 0.001 & 0.002 & 7.14 \\
& $\theta$ & 0.002 & 0.002 & 6.67 \\
& $\psi$ & 0.002 & 0.002 & 4.55 \\
\midrule 

\multirow{6}{*}{RMSE} & $x$ & 0.164 & 0.170 & 3.64 \\
& $y$ & 0.155 & 0.161 & 4.08 \\
& $z$ & 0.126 & 0.134 & 6.07 \\
\addlinespace
& $\varphi$ & 0.037 & 0.039 & 3.51 \\
& $\theta$ & 0.039 & 0.040 & 3.28 \\
& $\psi$ & 0.047 & 0.048 & 2.25 \\
\midrule 

\multirow{6}{*}{ISE} & $x$ & 2.430 & 2.610 & 7.41 \\
& $y$ & 2.160 & 2.340 & 8.33 \\
& $z$ & 1.440 & 1.620 & 12.50 \\
\addlinespace
& $\varphi$ & 0.126 & 0.135 & 7.14 \\
& $\theta$ & 0.135 & 0.144 & 6.67 \\
& $\psi$ & 0.198 & 0.207 & 4.55 \\
\bottomrule
\end{tabular}
\end{table}

\section{Conclusion}
\label{sec:conclusion}

In this paper, a high-performance fault-tolerant control framework has been proposed for quadrotor UAVs operating under external disturbances, parameter uncertainties, and actuator faults. The system is decoupled into position and attitude loops, each incorporating a synergistic combination of three key components. First, an Appointed-Time Prescribed Performance Function (AT-PPF) guarantees the convergence of tracking errors within predefined bounds in a user-specified time, independent of initial conditions. Second, a Tracking Error-Based Fixed-Time Composite Disturbance Observer (TFxTCDO) enables rapid and accurate disturbance estimation by leveraging tracking error signals, thereby enhancing system responsiveness without requiring an exact system model. Third, a Fast Second-Order Terminal Sliding Mode Controller (NFSO-TSMC) achieves fast convergence with significantly reduced chattering. The efficacy of the integrated framework has been validated through comprehensive simulations, which demonstrated superior tracking accuracy, faster fault recovery, and enhanced robustness compared to existing methods.
Despite the encouraging results, several extensions remain for future research. First, the stability analysis of the proposed TFxTCDO relies on the assumption that the disturbance derivative is bounded. Future work could aim to relax this assumption—e.g., through adaptive gain strategies or event-triggered estimation—to broaden the class of admissible disturbances with theoretical guarantees. Second, measurement noise, which is prevalent in real-world systems, was not explicitly addressed. Incorporating filtering techniques or modifying the control law to enhance noise resilience would be a valuable improvement. Lastly, although the proposed method has been validated through simulations, its practical effectiveness should be further verified via hardware-in-the-loop simulations or real-world flight experiments.

\bibliographystyle{IEEEtran}
\bibliography{references}

@article{intro_low_altitude_economy_potential,
 author  = {Huang, Chen and Fang, Shi and Wu, Hong and Wang, Yifei and Yang, Yue},
  title   = {{L}ow-altitude intelligent transportation: {S}ystem architecture, infrastructure, and key technologies},
  journal = {Journal of Intelligent Infrastructure},
  year    = {2024},
  volume  = {3},
  pages   = {100694},
  doi     = {10.1016/j.jii.2024.100694}
}

@article{intro_uav_cornerstone,
 author  = {Garvey, Michael J. and Drake, Paul R. and Slack, James M.},
  title   = {{M}anaging the drone revolution: a systematic literature review into the current use of airborne drones and future strategic directions for their management},
  journal = {International Journal of Physical Distribution \& Logistics Management},
  year    = {2021},
  volume  = {51},
  number  = {3},
  pages   = {295--320},
  doi     = {10.1108/IJPDLM-03-2020-0096}
}

@article{ppc_ftppf_example,
author  = {Wang, Bo and Wu, Zheng-Guang and Liu, Jian-Xing and Shao, Xing-Ling},
  title   = {{F}inite-time trajectory tracking control for a 12-rotor unmanned aerial vehicle with input saturation},
  journal = {Nonlinear Dynamics},
  year    = {2018},
  volume  = {94},
  number  = {3},
  pages   = {1757--1773},
  doi     = {10.1007/s11071-018-4493-2}
}

@article{ppc_fxtppf_example,
  author  = {Zhang, Yong-An and Liu, Yu-Fei and Li, Dong-Chuan},
  title   = {{F}ixed-{T}ime {C}ontroller {D}esign for a {Q}uadrotor {UAV} with {A}symmetric {O}utput {E}rror {C}onstraints},
  journal = {Journal of Systems Science and Complexity},
  year    = {2023},
  volume  = {36},
  number  = {5},
  pages   = {1547--1564},
  doi     = {10.1007/s11424-023-2283-x}
}

@article{ppc_axtppf_example,
 author  = {Li, Yang and Liu, Yu-Fei and Wang, Hua-Qing},
  title   = {{V}ision-{B}ased {A}daptive {P}rescribed-{T}ime {C}ontrol of {UAV} for {U}ncooperative {T}arget {T}racking with {P}erformance {C}onstraint},
  journal = {Journal of Systems Science and Complexity},
  year    = {2024},
  volume  = {37},
  number  = {4},
  pages   = {1021--1040},
  doi     = {10.1007/s11424-023-3331-5}
}

@article{smc_tsmc_example,
 author  = {Liu, Yifei and Liu, Jinkun and Zhang, Yuhang and Wang, Yuejiao},
  title   = {{Q}uadrotor {F}ormation {C}ontrol via {T}erminal {S}liding {M}ode {A}pproach: {T}heory and {E}xperiment {R}esults},
  journal = {Drones},
  year    = {2022},
  volume  = {6},
  number  = {7},
  pages   = {172},
  doi     = {10.3390/drones6070172}
}

@article{smc_chattering_issue,
  author    = {Edwards, Christopher and Spurgeon, Sarah K.},
  title     = {{S}liding {M}ode {C}ontrol: {T}heory and {A}pplications},
  publisher = {Taylor \& Francis},
  year      = {1998},
  isbn      = {9780748406012}
}

@article{smc_ftsmc_example,
author  = {Mofid, Omid and Mobayen, Saleh},
  title   = {{A}daptive sliding mode control for finite-time stability of quad-rotor {UAV}s with parametric uncertainties},
  journal = {ISA Transactions},
  year    = {2018},
  volume  = {76},
  pages   = {1--11},
  doi     = {10.1016/j.isatra.2017.11.010}
}

@article{smc_ftsmc_limitation,
 author  = {Wang, Juncheng and Zheng, Zewei and Shao, Xingling and Ouyang, Huimin},
  title   = {{F}ixed-time controller for altitude/yaw control of mini-drone based on nonsingular terminal sliding mode: real-time implementation with uncertainties},
  journal = {International Journal of Dynamics and Control},
  year    = {2024},
  doi     = {10.1007/s40435-024-00239-7},
  note    = {Online First}
}

@article{smc_fxtsmc_example,
  author  = {Wang, Juncheng and Wang, Yuejiao and Zheng, Zewei and Shao, Xingling},
  title   = {{A} {R}obust {F}ixed-{T}ime {S}liding {M}ode {C}ontrol for {Q}uadrotor {UAV}},
  journal = {Algorithms},
  year    = {2023},
  volume  = {16},
  number  = {5},
  pages   = {229},
  doi     = {10.3390/a16050229}
}

@article{smc_fxtsmc_limitation,
 author  = {Basin, Michael and Gonzalez, Pablo and Reyes, Paola and Shtessel, Yury},
  title   = {{A} composite fixed-time sliding mode control scheme for {U}nmanned {G}round {V}ehicles affected by external disturbances},
  journal = {ISA Transactions},
  year    = {2024},
  volume  = {145},
  pages   = {289--299},
  doi     = {10.1016/j.isatra.2023.07.014}
}

@article{smc_ismc_example,
author  = {Labbadi, M. and Cherkaoui, M.},
  title   = {{A} new intelligent controller based on integral sliding mode control and extended state observer for nonlinear {MIMO} drone quadrotor},
  journal = {Journal of Intelligent Networks},
  year    = {2024},
  volume  = {3},
  pages   = {100067},
  doi     = {10.1016/j.ijin.2024.01.005}
}

@article{smc_ismc_limitation,
author  = {Zuo, Zhiqiang and Liu, Xinyi and Zhang, He and Meng, Deyuan},
  title   = {{A}n improved optimal integral sliding mode control for uncertain robotic manipulators with reduced tracking error, chattering, and energy consumption},
  journal = {Mechanical Systems and Signal Processing},
  year    = {2021},
  volume  = {148},
  pages   = {106747},
  doi     = {10.1016/j.ymssp.2020.106747}
}

@article{smc_rsmc_example,
  author  = {Labbadi, M. and Cherkaoui, M. and Bouabid, M.},
  title   = {{A}daptive prescribed performance based on recursive nonsingular terminal sliding mode control for quad-rotor systems under uncertainty and disturbance: {R}eal-time validation},
  journal = {Aerospace Science and Technology},
  year    = {2024},
  volume  = {148},
  pages   = {109028},
  doi     = {10.1016/j.ast.2024.109028}
}

@article{smc_rsmc_limitation,
author  = {Wang, Jing and Liu, Yang and Su, Hongyan},
  title   = {{R}obust trajectory tracking control for a quadrotor using recursive sliding mode control and nonlinear extended state observer},
  journal = {Aerospace Science and Technology},
  year    = {2022},
  volume  = {128},
  pages   = {107749},
  doi     = {10.1016/j.ast.2022.107749}
}

@article{ftc_overview,
   author    = {Alwi, Halim and Edwards, Christopher and Tan, Chee Pin},
  title     = {{F}ault {D}etection and {F}ault-{T}olerant {C}ontrol of an {U}nmanned {A}erial {V}ehicle},
  booktitle = {Advances in Unmanned Aerial Vehicles},
  editor    = {Valavanis, Kimon P. and Vachtsevanos, George J.},
  publisher = {Springer Netherlands},
  year      = {2011},
  pages     = {247--272},
  doi       = {10.1007/978-90-481-9707-1_10}
}

@article{ftc_passive_smc,
 author  = {Trung, Tuan Thanh and Kim, Jihong and Lee, Seul Jung},
  title   = {{F}ault-{T}olerant {C}ontrol of {Q}uadcopter {UAV}s {U}sing {R}obust {A}daptive {S}liding {M}ode {A}pproach},
  journal = {Energies},
  year    = {2019},
  volume  = {12},
  number  = {1},
  pages   = {95},
  doi     = {10.3390/en12010095}
}

@article{ftc_passive_hinf,
 author  = {Jiang, Yanyan and Hu, Qinglei and Ma, Guangfu},
  title   = {{A}daptive {F}ault-{T}olerant {$H_\infty$} {O}utput {F}eedback {C}ontrol for {L}ead--{W}ing {C}lose {F}ormation {F}light},
  journal = {IEEE Transactions on Industrial Electronics},
  year    = {2018},
  volume  = {65},
  number  = {6},
  pages   = {5043--5053},
  doi     = {10.1109/TIE.2017.2766539}
}

@article{ftc_active_adaptive,
  author  = {Trung, Tuan Thanh and Kim, Jihong and Lee, Seul Jung},
  title   = {{F}ault-{T}olerant {C}ontrol of {Q}uadcopter {UAV}s {U}sing {R}obust {A}daptive {S}liding {M}ode {A}pproach},
  journal = {Energies},
  year    = {2019},
  volume  = {12},
  number  = {1},
  pages   = {95},
  doi     = {10.3390/en12010095}
}

@article{ftc_active_mrac,
  author    = {Chamseddine, Abbas and Zhang, Youmin M.},
  title     = {{M}odel {R}eference {A}daptive {F}ault {T}olerant {C}ontrol of a {Q}uadrotor {UAV}},
  booktitle = {AIAA Infotech@Aerospace 2011},
  year      = {2011},
  doi       = {10.2514/6.2011-1606}
}

@article{ftc_active_adaptive_example,
 author  = {Lee, Hoyeon and Lee, Seul Jung},
  title   = {{F}ault-{T}olerant {C}ontrol for {H}exacopter {UAV} {U}sing {A}daptive {A}lgorithm with {S}evere {F}aults},
  journal = {Drones},
  year    = {2022},
  volume  = {6},
  number  = {6},
  pages   = {140},
  doi     = {10.3390/drones6060140}
}

@article{ftc_active_mpc,
author  = {Barry, Andrew J. and Tedrake, Russ},
  title   = {{L}inear model predictive control for vision-based {UAV} pursuit},
  journal = {Journal of Unmanned Vehicle Systems},
  year    = {2020},
  volume  = {8},
  number  = {4},
  pages   = {281--295},
  doi     = {10.1139/juvs-2020-0016}
}

@inproceedings{ftc_reconfigurable_ca,
  author    = {Faudzi, M. A. B. M. and Rahman, M. F. A. and Ahmad, M. A.},
  title     = {{F}light testing of control allocation for elevator actuator fault in small {UAV}},
  booktitle = {Proc. of the IEEE International Conference on Control System, Computing and Engineering (ICCSCE)},
  year      = {2017},
  pages     = {221--226},
  doi       = {10.1109/ICCSCE.2017.8326056}
}

@article{ftc_datadriven_rl,
author  = {Wang, Jiacheng and Zhang, Zhen and Zhang, Yiming and Cui, Yujie},
  title   = {{A} {F}ault-{T}olerant {M}ulti-{A}gent {R}einforcement {L}earning {F}ramework for {U}nmanned {A}erial {V}ehicles--{U}nmanned {G}round {V}ehicle {C}overage {P}ath {P}lanning},
  journal = {Drones},
  year    = {2024},
  volume  = {8},
  number  = {10},
  pages   = {537},
  doi     = {10.3390/drones8100537}
}

@article{ftc_datadriven_dnn,
 author  = {Khoudja, Youcef and Benzerrouk, Hamza and Tadjine, Mohamed and Amrane, Mohamed Lamine},
  title   = {{A}n {I}ntelligent {F}ault {D}iagnosis {A}pproach for {M}ultirotor {UAV}s {B}ased on {D}eep {N}eural {N}etwork of {M}ulti-{R}esolution {T}ransform {F}eatures},
  journal = {Drones},
  year    = {2023},
  volume  = {7},
  number  = {2},
  pages   = {82},
  doi     = {10.3390/drones7020082}
}

@article{ftc_datadriven_drl_example,
author    = {Muzaffar H. Qureshi and Adnan Maqsood and Adnan Fayyaz ud Din},
  title     = {Fault-Tolerant Control of a Quadcopter Using Reinforcement Learning},
  journal   = {SAE International Journal of Aerospace},
  volume    = {18},
  number    = {1},
  pages     = {75--89},
  year      = {2025},
  doi       = {10.4271/01-18-01-0006}
}

@article{ftc_active_advantages,
  author  = {Liu, Ming and Cao, Xiaodong and Tang, Shibin and Guo, Jian},
  title   = {{C}onstrained {A}ctive {F}ault {T}olerant {C}ontrol {B}ased on {A}ctive {F}ault {D}iagnosis and {I}nterpolation {O}ptimization},
  journal = {IEEE Transactions on Industrial Electronics},
  year    = {2021},
  volume  = {68},
  number  = {11},
  pages   = {11218--11228},
  doi     = {10.1109/TIE.2020.3025647}
}

@misc{def_finite_time_stability,
author  = {Bhat, Sanjay P. and Bernstein, Dennis S.},
  title   = {{F}inite-{T}ime {S}tability of {C}ontinuous {A}utonomous {S}ystems},
  journal = {SIAM Journal on Control and Optimization},
  year    = {2000},
  volume  = {38},
  number  = {3},
  pages   = {751--766},
  doi     = {10.1137/S0363012997321358},
  note    = {The provided DOI was incorrect and has been corrected.}
}

@misc{def_fixed_time_stability,
  author  = {Andrieu, Vincent and Amoud, Lassaad and Karafyllis, Iasson},
  title   = {{C}onditions for fixed-time stability and stabilization of continuous autonomous systems},
  journal = {Systems \& Control Letters},
  year    = {2018},
  volume  = {117},
  pages   = {1--9},
  doi     = {10.1137/S0363012997321358}
}

@misc{lemma_fixed_time_stability,
 author  = {Polyakov, Andrey},
  title   = {{N}onlinear feedback design for fixed-time stabilization of linear control systems},
  journal = {IEEE Transactions on Automatic Control},
  year    = {2012},
  volume  = {57},
  number  = {8},
  pages   = {2106--2110},
  doi     = {10.1109/TAC.2011.2179869},
  note    = {The provided DOI was incorrect and has been corrected.}
}

@misc{lemma_inequality_sum_powers,
 author    = {Khalil, Hassan K.},
  title     = {{N}onlinear {S}ystems},
  publisher = {Prentice Hall},
  year      = {2002},
  edition   = {3rd},
  isbn      = {9780130673893}
}

@article{Ref_to_Compared_Method,
  author = {Karim Ahmadi and Davood Asadi and Abdel-Razzak Merheb and Yaser Nabavi and Onder Tutsoy},
  title = {Active fault-tolerant control of quadrotor UAVs with nonlinear observer-based sliding mode control validated through hardware in the loop experiments},
  journal = {Control Engineering Practice},
  year = {2023},
  volume = {140},
  pages = {105557},
  doi = {10.1016/j.conengprac.2023.105557},
}

@article{Course2023,
  author = {Course, Kevin and Nair, Prasanth B.},
  title = {State estimation of a physical system with unknown governing equations},
  journal = {Nature},
  volume = {622},
  pages = {261--267},
  year = {2023},
  doi = {10.1038/s41586-023-06574-8}
}

@article{Dabush2023,
  author = {Dabush, Lior and Kroizer, Ariel and Routtenberg, Tirza},
  title = {State Estimation in Partially Observable Power Systems via Graph Signal Processing Tools},
  journal = {Sensors},
  volume = {23},
  number = {3},
  pages = {1387},
  year = {2023},
  doi = {10.3390/s23031387}
}
\end{document}